\documentclass{article}

\usepackage{microtype}
\usepackage{graphicx}
\usepackage{subcaption}
\usepackage{caption}
\usepackage{booktabs}
\usepackage{hyperref}
\usepackage{amsmath}
\usepackage{amssymb}
\usepackage{mathtools}
\usepackage{amsthm}
\usepackage[capitalize,noabbrev]{cleveref}
\usepackage{xcolor}
\usepackage[normalem]{ulem}
\usepackage{bm}
\usepackage{placeins}
\usepackage{xurl}
\usepackage{balance}
\usepackage{float}

\usepackage[accepted]{icml2026}
\makeatletter
\renewcommand{\Notice@String}{\textit{Accepted to the 2nd Workshop on Compositional Learning at ICML 2026, Seoul, South Korea. Copyright 2026 by the author(s).}}
\makeatother
\icmltitlerunning{Dissociating Decodability and Causal Use in Bracket-Sequence Transformers}

\begin{document}

\twocolumn[
\icmltitle{Dissociating Decodability and Causal Use in Bracket-Sequence Transformers}

\icmlsetsymbol{equal}{*}

\begin{icmlauthorlist}
\icmlauthor{Aryan Sharma}{yale}
\icmlauthor{Cutter Dawes}{ind}
\icmlauthor{Shivam Raval}{harvard}
\end{icmlauthorlist}

\icmlaffiliation{yale}{Yale University}
\icmlaffiliation{ind}{Independent}
\icmlaffiliation{harvard}{Harvard University}

\icmlcorrespondingauthor{Aryan Sharma}{aryan.sharma@yale.edu}

\icmlkeywords{mechanistic interpretability, probes, transformers, Dyck language}

\vskip 0.3in
]

\printAffiliationsAndNotice{}

\begin{abstract}
When trained on tasks requiring an understanding of hierarchical structure, transformers have been found to represent this hierarchy in distinct ways: in the geometry of the residual stream, and in stack-like attention patterns maintaining a last-in, first-out ordering. However, it remains unclear whether these representations are causally used or merely decodable. We examine this gap in transformers trained on the Dyck language (a formal language of balanced bracket sequences), where the hierarchical ground truth is explicit. By probing and intervening on the residual stream and attention patterns, we find that depth, distance, and top-of-stack signals are all decodable, yet their causal roles diverge. Specifically, masking attention to the true top-of-stack position causes a sharp drop in long-distance accuracy, while ablating the low-dimensional residual stream subspaces that encode this hierarchical information has comparatively little effect. This effect extends even to natural language settings, suggesting that even in a controlled setting where the relevant hierarchical variables are known, decodability alone does not imply causal use.
\end{abstract}

\begin{figure*}[!t]
\centering
\includegraphics[width=\textwidth]{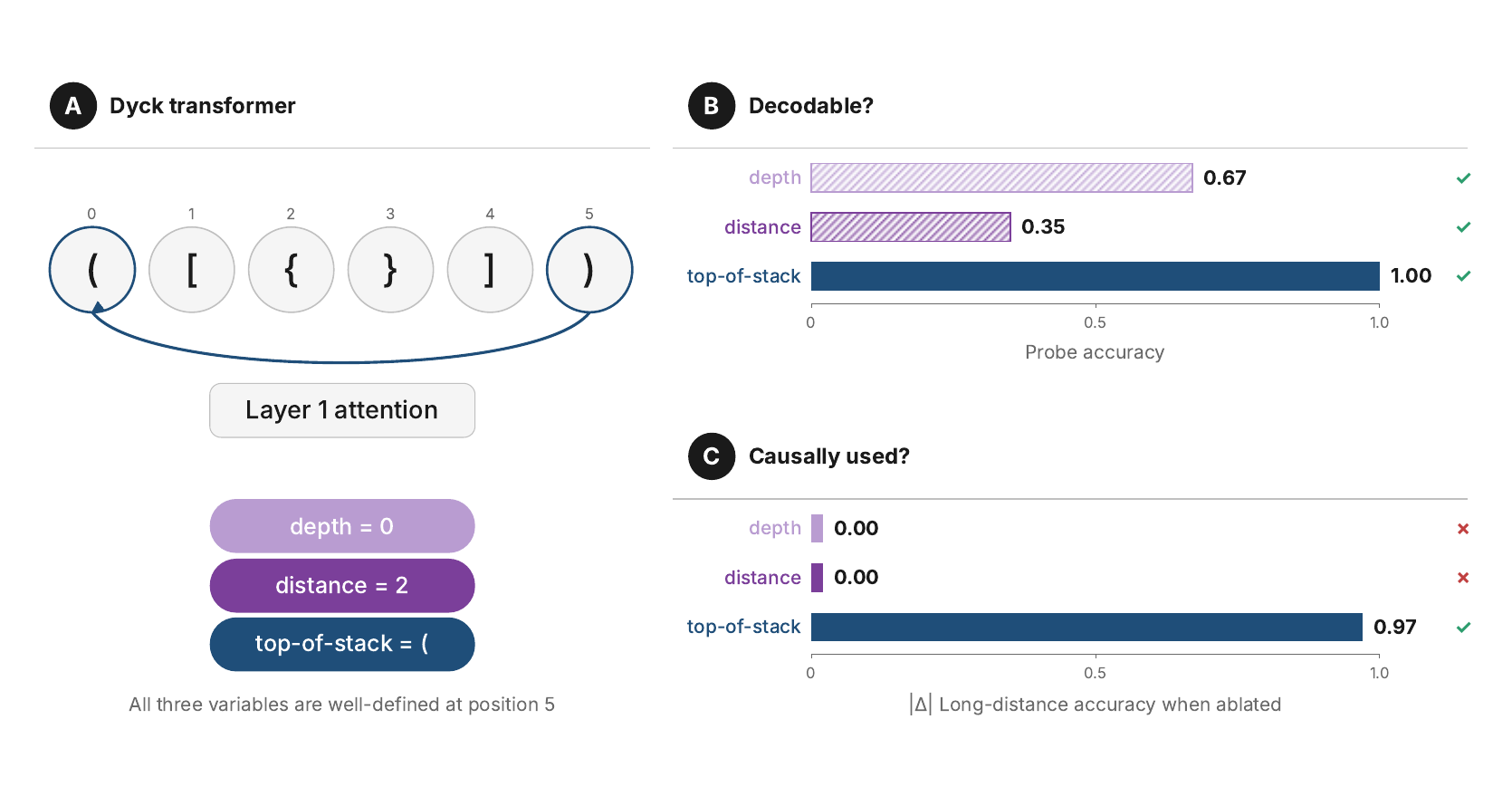}
\caption{\textbf{Decodability $\neq$ causal use.} \textbf{(A)} At a closing bracket in a trained Dyck transformer, layer-1 attention routes sharply to the matching opener, while three latent variables (depth, distance, and top-of-stack) are well-defined at this position. \textbf{(B)} All three variables are decodable from the model. \textbf{(C)} Only top-of-stack is causally used: ablating the probe-aligned residual-stream subspaces for depth and distance has near-zero effect on task performance, while masking the top-of-stack attention edge collapses it.}
\label{fig:teaser}
\end{figure*}

\section{Introduction}
Linear probes are widely used to test whether model states encode algorithmic and linguistic variables \citep{hewitt2019structural,hewitt2019control,elazar2021amnesic}. However, encoding does not directly imply use, and decodable variables have previously been seen to be behaviorally irrelevant \citep{hewitt2019control,elazar2021amnesic}. This gap has become increasingly important as many empirical claims in interpretability methodology rely on decodability alone. More broadly, false positives from multiple testing and underconstrained analysis pipelines are a familiar methodological risk, meaning this concern is not unique to interpretability \citep{meloux2025deadsalmon}.

We study this gap in the Dyck language setting---the formal language of balanced bracket sequences---where the relevant hierarchical variables are explicitly computable. Intuitively, a Dyck string is a sequence of nested brackets that must be closed in a last-in-first-out manner, so predicting each closing bracket requires tracking which opener is currently unmatched. This minimal, fully specified instance of hierarchical structure provides us with a controlled setting in which representational and causal claims can be separated cleanly, since the relevant variables can be probed directly and interventions can be evaluated against ground-truth structure. Specifically, we ask: when hierarchical variables are decodable within Dyck transformers, which are actually relevant to model computation?

Our main claim is that the behavior of these Dyck transformers depends more strongly on sparse top-of-stack attention routing (i.e., each closing bracket attending almost entirely to the current stack top) than on decoded residual-stream directions. We support this claim with five findings: 
\begin{enumerate}

    \item Hierarchical signals (depth, distance, and top-of-stack identity) are strongly decodable early in training.
    \item Depth and distance remain decodable OOD, yet ablating their probe-aligned subspaces has near-zero behavioral effect.
    \item Masking the true top-of-stack attention edge is catastrophic.
    \item Activation patching localizes the critical retrieval step to a single computational block.
    \item A templated natural-language setting shows the same dissociation outside of the Dyck language case.
\end{enumerate}  

\section{Related Work}
Prior work has shown that syntactic and hierarchical structure is linearly decodable from language model representations. In particular, \citet{hewitt2019structural} demonstrated that parse-tree distance and depth are linearly decodable from transformer hidden states, and we test whether they are causally used. Subspace-removal methods extend probing by ablating decoded directions to test causal use \citep{ravfogel2020null,ravfogel2022rlace}, and have found that certain decoded features can be removed without affecting downstream behavior, which motivates our decodability-versus-causality framing in a controlled setting.

Related work in causal abstraction \citep{geiger2021causal} and mechanistic interpretability \citep{elhage2021framework,conmy2023acdc,wang2022ioi} emphasizes specific attention pathways rather than globally decodable geometry. Notably, \citet{wang2022ioi} identified a small circuit of attention edges carrying the indirect-object-identification task in GPT-2. Our attention-knockout finding is a formal-language instance of this pattern, with the advantage that the ground-truth mechanism is explicitly computable.

The Dyck language is a standard testbed for hierarchical structure in transformers \citep{yao2021selfattention,murty2023grokking,tiwari2025emergent,hayakawa2025theoretical,wen2023transformers}. We add a causal, per-model analysis differentiating linearly decodable features from those used at inference time.
\section{Experimental setup}

\paragraph{Task and models.}
The Dyck-$(k,m)$ language consists of balanced strings with $k$ bracket types and maximum depth $m$, where depth counts the number of open brackets. For example, in \texttt{([\{\}])}, the depth after each token is $(1, 2, 3, 2, 1, 0)$. We take distance to be the parse-tree distance between token pairs, following the methodology of the structural probe in \citet{hewitt2019structural}; in the example above, the closing \texttt{)} and its matching opener \texttt{(} are at parse-tree distance $2$ (Figure~\ref{fig:teaser}). At each position, we refer to the \textit{stack top} as the most recently opened unclosed bracket. We use the Dyck language because it exposes explicit latent hierarchical variables---including stack top, depth, and distance---in a setting where both probing and intervention can be evaluated against the ground truth.

As a baseline setting, we train on Dyck-$(20,10)$ with 2-layer, 1-head transformers, embedding dimension $d \in \{16,32,64\}$, $5 \times 10^5$ optimization steps, a training set of $10^4$ sequences, and a held-out test set of $10^3$ sequences with no overlap. Data generation details are in Appendix~\ref{app:data}. On held-out IID evaluation data, these base models achieve 98.2\%--100\% long-distance accuracy (i.e., accuracy on matched bracket pairs whose closing position is at least 10 positions from the corresponding opening position) across the three widths. Architecture and training details are in Appendix~\ref{app:training}.

We additionally scale task difficulty by varying $(k,m)\in\{(20,18),(30,14),(40,10),(40,18)\}$ at $d = 64$. Training details and full results are in Appendix~\ref{app:harder}.\footnote{All code is available at \url{https://github.com/aryans-15/bracket-transformers}.}

\section{Probing decodability of task-relevant quantities}
\label{sec:decodability}
\paragraph{Residual-stream and attention probes.}
We probe depth and distance from the layer-2 residual stream and report the Pearson correlation between predicted and true values as our primary metric. Probe architectures, training details, and R$^2$/MSE numbers are in Appendices~\ref{app:probes} and~\ref{app:probe-metrics}. We additionally sweep over probe capacity to show that regularization does not change the qualitative OOD pattern (Appendix~\ref{app:probe-sweep}).

When probing top-of-stack position, we read it directly from the attention pattern, since the model's prediction at a closing bracket depends on routing attention to a single source token. Specifically, at a closing-bracket query $q$ with ground-truth stack-top source position $s(q)$, we report the \emph{attention miss} as  $1 - a^{(1)}_{q,\,s(q)}$, where $a^{(1)}_{q,\cdot}$ is the layer-1 attention row at $q$. This has no fitted probe parameters, so overfitting is not a concern. A logistic-regression probe on the residual stream is reported in Appendix~\ref{app:residual-tos-probe}.

We note that the top-of-stack identity and distance are related but not redundant, since top-of-stack fixes which bracket type is the next closer (categorical), while distance measures how far back the corresponding opener sits (scalar). Knowing distance at a closing bracket determines identity via the sequence, but the converse does not hold. Meanwhile, depth is orthogonal to both.

\begin{figure}[t]
    \centering
    \includegraphics[width=\linewidth]{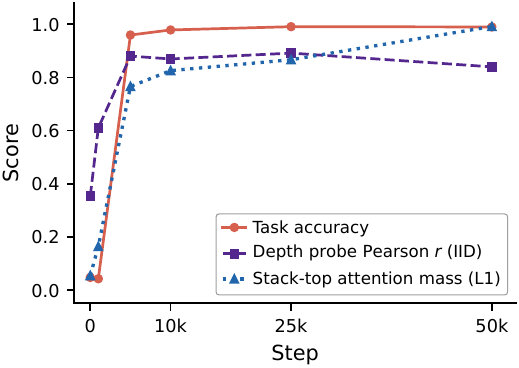}
    \caption{Training dynamics for the $d=64$ base model on the first 50k steps. Task accuracy (red), depth probe Pearson IID (dark purple), and layer-1 stack-top attention mass (blue) all stabilize by step~5k. The full 500k-step trajectory is in Figure~\ref{fig:training_dynamics_log}.}
    \label{fig:training_dynamics}
\end{figure}

\paragraph{Hierarchical signals emerge early.}
Figure~\ref{fig:training_dynamics} shows all three signals over training. By step 5k, top-of-stack attention mass and depth probe Pearson correlation are already strong, and task accuracy is near ceiling. This indicates that depth, distance, and top-of-stack identity are recoverable before complete task convergence. Probe accuracy as a function of sequence position is in Appendix~\ref{app:probe-position}.

\begin{figure}[t]
    \centering
    \includegraphics[width=\linewidth]{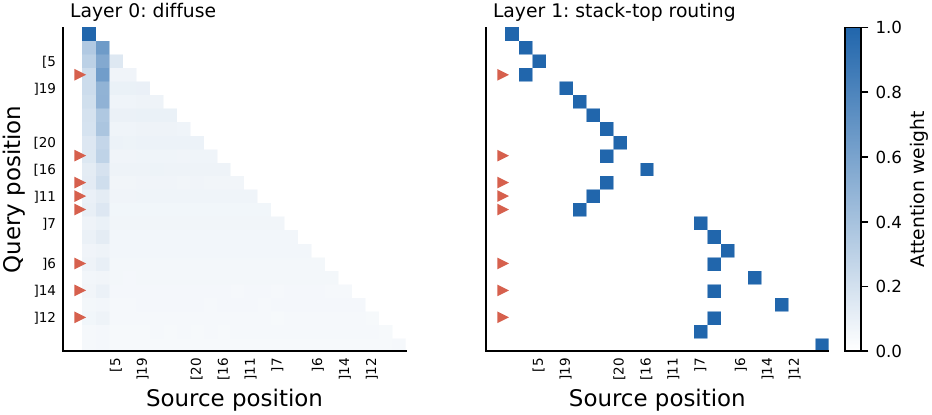}
    \caption{Attention weights for a representative length-24 sequence. Orange triangles mark closing-bracket query rows. Layer 0 (left) attention is diffuse; Layer 1 (right) collapses each marked row onto the matching opener's column.}
    \label{fig:attention_heatmap}
\end{figure}

Figure~\ref{fig:attention_heatmap} shows the corresponding picture through attention. Each row of the heatmap is an attention distribution for a single query position, and the orange triangles mark the rows corresponding to closing brackets, i.e., the positions whose next-token prediction depends on retrieving the matching opener. In layer~0, these marked rows are diffuse, meaning attention is spread broadly across source positions. In layer~1, each marked row collapses onto a single bright cell, which is consistently the column of the matching opening bracket. We show this mechanism is causally necessary in Section~\ref{sec:causality}.

\paragraph{Out-of-distribution decodability.}
Figure~\ref{fig:ood_bars} reports decodability across IID, OOD-length, and OOD-depth splits (defined in Appendix~\ref{app:data}). The top-of-stack attention is near-perfect on all three splits, while the depth and distance probe Pearson correlations degrade but stay positive. The positive Pearson alongside negative OOD $R^2$ (Appendix~\ref{app:probe-metrics}) indicates the probes preserve the rank ordering of depth and distance OOD but are poor absolute predictors, whereas top-of-stack stays robustly recoverable. We additionally sweep probe capacity to reproduce the same qualitative pattern, verifying that OOD degradation is not a probe-overfitting artifact (Appendix~\ref{app:probe-sweep}).

\begin{figure}[t]
    \centering
    \includegraphics[width=\linewidth]{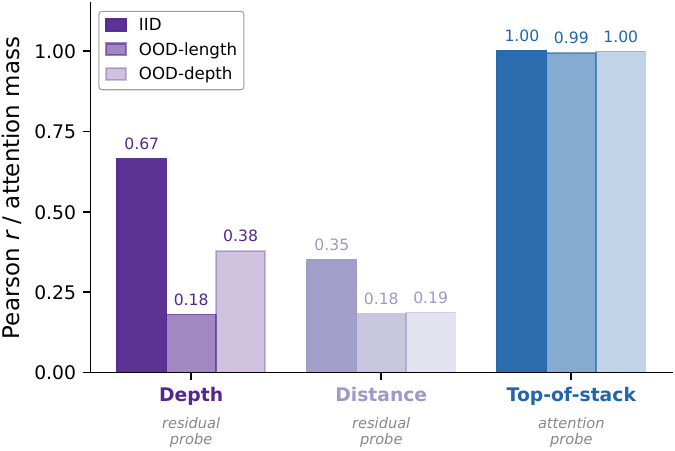}
    \caption{Decodability across splits. Depth and distance (purple) report residual-probe Pearson $r$; top-of-stack (blue) reports layer-1 attention mass. Top-of-stack stays near-perfect across IID, OOD-length, and OOD-depth; depth and distance degrade but stay positive. Long-distance task accuracy is $\geq 0.994$ across all splits.}
    \label{fig:ood_bars}
\end{figure}

\section{Interventions identify causally important quantities}
\label{sec:causality}

\paragraph{Causal interventions.}
For depth and distance, we ablate probe-identified subspaces in the residual stream at ranks $r \in \{2, 4, 8, 16, 32, 64\}$, comparing against random rank-matched baselines (further details in Appendix~\ref{app:causal-details}, and intervention strength sensitivity analysis in Appendix~\ref{app:alpha-sweep}). For top-of-stack, we perform attention knockout by zeroing the attention score to the true source position before softmax \citep{wang2022ioi}, against random control edges. Query/key-space interventions are in Appendix~\ref{app:qk-interventions}.

\paragraph{Activation patching.}
To localize \textit{where} in computation the stack-top signal becomes sufficient, we use activation patching \citep{meng2022rome}: we substitute clean activations at each sublayer checkpoint of a corrupted run and measure recovery. Setup details, including corruption strategy, cache points, and offset sweep, are in Appendix~\ref{app:patching-setup}.

\paragraph{Depth and distance are not causally necessary.}
Ablating the decoded depth and distance directions from the residual stream has little effect. Depth ablation produces zero drop on the harder variants despite a final probe Pearson $r$ of $+0.67$ on IID, indicating the model encodes depth but does not use it for prediction. Similarly, ablating probe-aligned distance subspaces at low ranks has a near-zero effect, while removing a random subspace of the same rank causes substantial drops (Appendix~\ref{app:distance-sweep}). These residual directions are not behaviorally necessary, but query/key-space interventions (Appendix~\ref{app:qk-interventions}) show they participate in computing layer-1 routing, meaning the decodable geometry is likely consumed by the attention mechanism.

\paragraph{Top-of-stack routing is causally necessary.}
In our base $d = 64$ models, layer~1 attention to the true stack-top position is $\approx 0.98$--$0.997$, compared to a uniform baseline of $\approx 0.013$ (Appendix~\ref{app:harder}). This routing mechanism is stable across seeds (Appendix~\ref{app:multiseed}).

\begin{figure}[t]
    \centering
    \includegraphics[width=\linewidth]{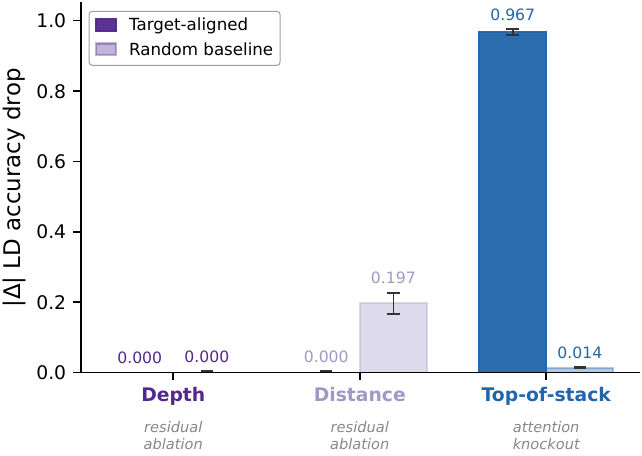}
   \caption{Causal intervention $|\Delta|$ long-distance accuracy: residual ablation of depth (left), residual ablation of rank-2 distance subspace (middle), and layer-1 attention-edge knockout of top-of-stack (right).}
    \label{fig:causal_bars}
\end{figure}

To test causal necessity, we perform attention knockout. We find that masking the true stack-top edge drops long-distance accuracy by $-0.967 \pm 0.009$ on the harder-variant test sets, while masking a randomly chosen edge only drops it by $-0.014 \pm 0.002$. Figure~\ref{fig:causal_bars} shows all three interventions on a common axis: residual-stream depth ablation, residual-stream distance ablation (at low rank), and attention-edge knockout of top-of-stack. Only the attention-edge knockout collapses accuracy.

\begin{figure}[t]
    \centering
    \includegraphics[width=\linewidth]{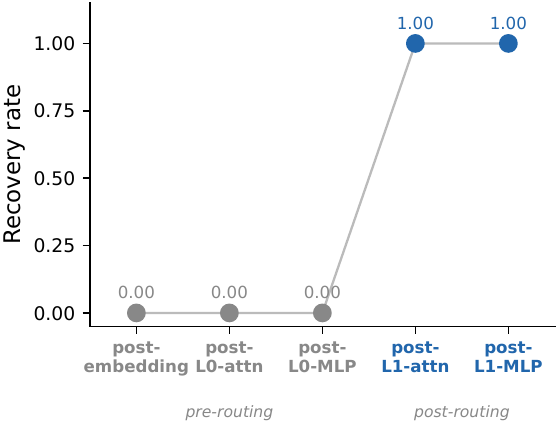}
    \caption{Activation patching recovery rate at the closing-bracket query position. Recovery jumps from 0 to 1 between post-L0-MLP and post-L1-attention. Full offset heatmap in Appendix~\ref{app:patching-full}.}
    \label{fig:patching}
\end{figure}

\paragraph{Activation patching localizes the mechanism to layer~1 attention.}
Our attention masking result shows the stack-top edge is necessary, but does not answer at what stage that signal becomes sufficient. To bridge this gap, we turn to activation patching. Figure~\ref{fig:patching} shows that patching before layer~1 attention yields essentially no recovery, while patching at or after layer~1 attention at the closing bracket position produces 100\% recovery. This indicates that the critical retrieval step is complete at layer-1 attention, and once that operation is correct, the residual state is sufficient for task success.

\paragraph{The dissociation persists in natural language.}
To test whether this dissociation extends beyond formal languages, we train a transformer with the same architecture ($d=64$, 2 layers, 1 head) on a templated subject--verb agreement task. In the task, each input has one subject plus 0--3 prepositional-phrase attractors with a randomly chosen number; the model must match a final \texttt{<verb>} token to the given subject. We choose this setup because our template removes the redundant agreement cues of unconstrained English, forcing retrieval of the subject's number through attention rather than a local heuristic. Over four seeds, we find that our model achieves 99.2--99.4\% accuracy uniformly across attractor counts, ruling out a positional shortcut. We provide the full setup and per-seed numbers in Appendix~\ref{app:nl-pilot}.

We further find that all three findings from our Dyck experiments recur. First, the subject-number probe stays at $\geq 0.99$ accuracy IID and OOD, while the attractor-count probe collapses OOD ($R^2 = -12.3$), just as depth and distance did. Second, masking the single subject-to-verb attention edge at layer~1 drops accuracy to
chance on three of the four seeds. This aligns with the solution multiplicity in our Dyck setting (Appendix~\ref{app:multiseed}). Third, activation patching localizes the computation to layer-1 attention (recovery $1.000$ after layer-1 attention vs.\ $0.000$ after the embedding), identical to the mechanism in our Dyck setting.

\section{Discussion}
We study the relationship between decodability and causal importance in transformer representations, within a controlled setting containing explicit hierarchy. We find depth and distance to be decodable early and positively correlated OOD, yet ablating their probe-aligned subspaces has near-zero effect. Top-of-stack identity behaves differently: its attention probe is lossless under distribution shift, and masking the single attention edge collapses task performance. Activation patching further establishes that the top-of-stack attention mechanism is not only causally necessary, but its recovery is sufficient for task success. These results make clear the importance of dissociating between decodability and causality in transformers and other neural networks. We also stress that our analysis goes beyond correlational diagnostics, since output-facing tools such as the logit lens \citep{nostalgebraist2020logitlens} can show a variable is readable at a layer, but only our ablations and attention knockouts establish that it is causally necessary for the prediction.

Our study has some limitations. First, all models here are 2-layer, 1-head transformers, and whether the same dissociation holds in scaled models is untested. Second, both the Dyck strings and our natural-language pilot (Section~\ref{sec:causality}) are retrieval-based tasks whose success is determined by a single matching token, which could potentially favor sparse attention routing over distributed computation. For this reason, tasks requiring the compositional integration of several latent variables could show a weaker dissociation. We leave both of these to future work.

\section*{Acknowledgments}
We thank the Supervised Program for Alignment Research (SPAR) for
providing compute and resources to carry out this work.

\balance
\bibliography{icml2026_workshop}
\bibliographystyle{icml2026}

\clearpage
\appendix
\section{Appendix}
\label{app:main}
\subsection{Training and architecture details}
\label{app:training}

\begin{table}[!ht]
\centering
\small
\setlength{\tabcolsep}{4pt}
\caption{Training hyperparameters (all models).}
\label{tab:hyperparams}
\begin{tabular}{ll}
\toprule
Parameter & Value \\
\midrule
Optimizer & AdamW \\
Learning rate & $3\times10^{-4}$ (constant) \\
Adam $\beta_1, \beta_2$ & $(0.9,\,0.999)$ \\
Weight decay & $0.0$ \\
Batch size & 128 sequences \\
Gradient clipping & 1.0 (max norm) \\
Base model steps & 500k \\
Harder variant steps & 300k \\
\bottomrule
\end{tabular}
\end{table}

\begin{table}[!ht]
\centering
\small
\setlength{\tabcolsep}{4pt}
\caption{Model architecture (all variants).}
\label{tab:arch}
\begin{tabular}{ll}
\toprule
Property & Value \\
\midrule
Layers & 2 \\
Heads & 1 \\
MLP hidden dim & $4d$ (ReLU) \\
Positional encoding & Learned \\
Layer norm & Post-LN \\
Attention mask & Causal \\
\bottomrule
\end{tabular}
\end{table}

\begin{table}[!ht]
\centering
\small
\setlength{\tabcolsep}{4pt}
\caption{Parameter counts and vocabulary sizes (vocab size $= 2k+2$).
Base models use $d\in\{16,32,64\}$ with $k=20$; harder variants fix
$d=64$ and vary $k$.}
\label{tab:params}
\begin{tabular}{cccc}
\toprule
$d$ & $k$ & Parameters & Vocab size \\
\midrule
16 & 20 & 15,424 & 42 \\
32 & 20 & 43,136 & 42 \\
64 & 20 & 135,424 & 42 \\
64 & 30 & 136,704 & 62 \\
64 & 40 & 137,984 & 82 \\
\bottomrule
\end{tabular}
\end{table}

\begin{figure}[H]
    \centering
    \includegraphics[width=0.8\linewidth]{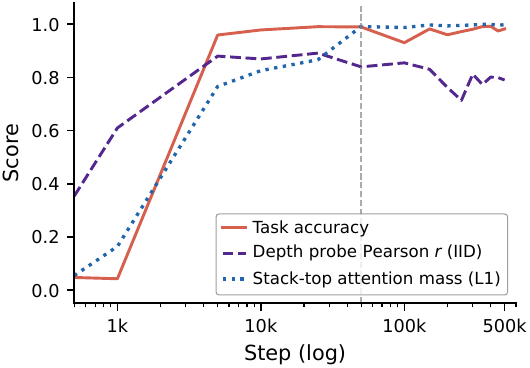}
    \caption{Full 500k-step training dynamics on a log x-axis. Task accuracy (red), depth probe Pearson IID (dark purple), and layer-1 stack-top attention mass (blue) all stabilize well before the end of training. Dashed vertical line marks step 50k.}
    \label{fig:training_dynamics_log}
\end{figure}

\paragraph{Full training-dynamics trajectory.}
Figure~\ref{fig:training_dynamics_log} extends the main-text Figure~\ref{fig:training_dynamics} to the full 500k-step trajectory on a log x-axis. All three signals stabilize well before the end of training. The dashed vertical line marks step 50k.

\subsection{Data generation and evaluation splits}
\label{app:data}

We sample Dyck strings with a stack-based probabilistic procedure. At each step, if the stack is empty, the model samples uniformly from $k$ opening brackets or $\textsc{end}$; if at maximum depth, it returns the closing bracket no matter what; and otherwise, it samples uniformly from $k$ opening brackets or the one correct closing bracket.

Our base model is trained on Dyck-$(20,10)$ with 10,000 training sequences (with length range of 22--512 and mean of 146.5) and 1,000 test sequences (with length range of 4--488 and mean of 51.9). The harder variants use 200,000 training and 20,000 test sequences each.

Out-of-distribution splits are constructed from our base set as follows. \emph{OOD-length}: we train on sequences with length $\leq 64$ (759 sequences) and evaluate on sequences with length $\geq 128$ (124 sequences). \emph{OOD-depth}: we train on sequences with max depth $\leq 4$ (567 sequences) and evaluate on sequences with max depth $\geq 8$ (288 sequences).

\subsection{Probe training details}
\label{app:probes}

All our probes are trained on an 80/20 split across a base 1,000-sequence set, with up to 128 tokens subsampled per
sequence, and hidden states extracted from the final transformer block without standardization. For activation patching experiments, we train probes on clean sequences only and apply them to corrupted forward passes.

Our depth probe is a ridge regressor with $\alpha=1.0$, our top-of-stack probe is a multinomial logistic regression (\texttt{lbfgs}, $C=1.0$), and our distance probe is a linear projection $B \in \mathbb{R}^{d \times 32}$ trained with AdamW (learning rate $3 \times 10^{-3}$, weight decay $10^{-4}$, cosine schedule) for 1500 steps, fitting $\lVert B h_i - B h_j \rVert_2$ to the true pairwise Dyck tree distance. We additionally ablate probe-aligned distance subspaces at ranks $r \in \{2, 4, 8, 16, 32, 64\}$ for the causal-intervention experiments (Appendix~\ref{app:distance-sweep}).

\subsection{Causal intervention details}
\label{app:causal-details}

\paragraph{Residual-stream ablation (depth).}
We remove probe-identified directions from the residual stream: if $U_r \in \mathbb{R}^{d \times r}$ is an orthonormal basis for a probe-aligned rank-$r$ subspace with projector $P_r = U_r U_r^\top$, we intervene on a hidden state $h$ via $h' = h - \alpha P_r h$, with $\alpha = 1$ in the main text (other strengths in Appendix~\ref{app:alpha-sweep}). As a baseline, we apply the same procedure to random rank-matched orthonormal subspaces in the same layer. For completeness, we apply the same residual-stream ablation procedure to the top-of-stack direction; results are in Appendix~\ref{app:stacktop-residual} and are consistent with our attention knockout findings.

\subsection{Activation patching setup}
\label{app:patching-setup}

We construct (clean, corrupted) sequence pairs by replacing a matched opening bracket with a wrong bracket type and only keeping pairs where the corrupted model incorrectly predicts the closing position. For each pair, we cache the residual stream at five sublayer boundaries in the clean run (post-embedding, post-attention, and post-MLP at each layer) and substitute the clean cache into the corrupted run one position and one cache point at a time, sweeping a $\pm 3$ token window around the closing bracket to account for variation in the opening bracket's location. Corruption strategy and probe tracking details are in Appendices~\ref{app:corruption} and~\ref{app:probe-corruption}.

\paragraph{Distance subspace ablation.}
For distance, we ablate the probe-aligned subspace at ranks $r \in \{2, 4, 8, 16, 32, 64\}$ and compare the behavioral effect against ablating a random rank-matched subspace. This tests whether the distance geometry encoded in the residual stream is causally used.

\paragraph{Attention knockout.}
For each query token $t$, we identify the source position corresponding to the true top-of-stack token and set the corresponding boolean attention-mask entry to \texttt{True} \citep{wang2022ioi}. This sends the relevant logit to $-\infty$ before softmax, thereby removing that edge from the attention graph. We compare masking the true top-of-stack edge against masking random control edges.

\subsection{Probe metrics: Pearson vs.\ R$^2$}
\label{app:probe-metrics}
The main text reports Pearson correlation between predicted and true depth/distance as the probe quality metric. For completeness, we also report R$^2$ and MSE in Table~\ref{tab:probe-metrics-ood}. We report Pearson because of scaling differences between IID and OOD splits; specifically, the mean depth shifts from $\approx 1$ IID to $\approx 5$ OOD-depth. These differences inflate MSE and R$^2$, making them inaccurate reflections of whether the probe recovers the underlying variable, while Pearson is scale-invariant.

\begin{table}[!ht]
\centering
\small
\setlength{\tabcolsep}{4pt}
\caption{Depth and distance probe metrics across splits ($d=64$ base model). Pearson is the main-text metric; R$^2$ and MSE are reported here for completeness.}
\label{tab:probe-metrics-ood}
\begin{tabular}{llccc}
\toprule
Probe & Split & Pearson $r$ & R$^2$ & MSE \\
\midrule
Depth     & IID        & $+0.67$ & $+0.44$ & $0.60$ \\
Depth     & OOD-length & $+0.18$ & $-2.20$ & $22.78$ \\
Depth     & OOD-depth  & $+0.38$ & $-1.00$ & $16.14$ \\
Distance  & IID        & $+0.35$ & $-0.17$ & $1.53$ \\
Distance  & OOD-length & $+0.18$ & $-1.98$ & $41.47$ \\
Distance  & OOD-depth  & $+0.19$ & $-1.18$ & $18.76$ \\
\bottomrule
\end{tabular}
\end{table}

\subsection{Probe capacity sweep}
\label{app:probe-sweep}
As a sensitivity check on probe capacity, we sweep the top-$k$ PC-regularized variants ($k \in \{8, 16, 32\}$) and full-rank Ridge $\alpha \in \{1, 10, 100, 1000\}$ for depth. We find that no configuration recovers positive R$^2$ on either OOD split, and the top-10 PC $h$-probe gives IID R$^2 = 0.08$, OOD-length $-3.25$, OOD-depth $-1.88$ --- worse than the full-rank $\alpha=1$ probe on IID and OOD alike, indicating the first 10 PCs of the layer-2 residual stream do not cleanly carry the depth signal. This qualitative collapse is robust across the sweep. The Pearson $r$ sweep shows the same qualitative pattern, as every probe-capacity configuration produces an OOD Pearson $r$ below IID, confirming that the main-text degradation is not an artifact of probe capacity.

\subsection{Residual stream top-of-stack probe}
\label{app:residual-tos-probe}
As a completeness check, we fit a multinomial logistic-regression probe for top-of-stack identity directly on the residual stream. This probe achieves $1.000$ accuracy across IID, OOD-length, and OOD-depth splits, consistent with the attention-based readout. The main text uses the attention pattern as the canonical stack-top readout because it has no fitted parameters and is therefore resistant to overfitting.

\subsection{Residual stream top-of-stack ablation}
\label{app:stacktop-residual}
As a completeness check, we ablate the residual-stream stack-top direction identified by the logistic-regression probe above. The ablation has a much smaller effect than the attention-edge knockout ($|\Delta|$ LD $\ll 0.967$), consistent with the finding that stack-top is causally used in the attention pattern rather than the residual stream.

\subsection{Harder variant results}
\label{app:harder}

\begin{table}[!ht]
\centering
\small
\setlength{\tabcolsep}{3pt}
\caption{Harder variant results on IID test sets. LD = long-distance accuracy, ToS = top-of-stack probe accuracy, Attn = mean layer~1 stack-top attention mass, Unif = uniform attention baseline, Edge $\Delta$ = change in LD acc.\ when stack-top attention edge is masked.}  
\label{tab:harder-full}
\begin{tabular}{lcccccc}
\toprule
Variant & LD acc. & Depth $R^2$ & ToS acc. & Attn & Unif & Edge $\Delta$ \\
\midrule
k20\_m18 & 0.9999 & 0.887 & 0.9999 & 0.986 & 0.013 & $-0.954$ \\
k30\_m14 & 0.9981 & 0.846 & 0.9999 & 0.983 & 0.014 & $-0.965$ \\
k40\_m10 & 0.9993 & 0.711 & 0.9999 & 0.990 & 0.017 & $-0.977$ \\
k40\_m18 & 0.9986 & 0.716 & 0.9999 & 0.983 & 0.013 & $-0.972$ \\
\bottomrule
\end{tabular}
\end{table}

\begin{table}[!ht]
\centering
\small
\setlength{\tabcolsep}{3pt}
\caption{Harder variant training summary. All runs use $d=64$, seed~42, AdamW lr=$3\times10^{-4}$. Steps to 99\% denotes the first checkpoint at which long-distance accuracy exceeds 0.99.}
\label{tab:training-harder}
\begin{tabular}{lccc}
\toprule
Variant & Final loss & Steps to 99\% & Wall-clock (h) \\
\midrule
k20\_m18 & 2.145 & 50k & 5.9 \\
k30\_m14 & 2.331 & 50k & 6.3 \\
k40\_m10 & 2.463 & 50k & 7.1 \\
k40\_m18 & 2.482 & 150k & 7.8 \\
\bottomrule
\end{tabular}
\end{table}

We report $R^2$ rather than Pearson $r$ in Table~\ref{tab:harder-full} because the harder variants are only evaluated IID, where the scale concerns motivating Pearson in the main text do not apply.

All four harder variants achieve a long-distance accuracy above 99.8\% and top-of-stack probe accuracy above 99.99\%. The depth $R^2$ decreases with $k$, indicating that depth becomes harder to linearly extract as bracket type complexity increases, yet the task performance still remains near ceiling, further emphasizing decodability without causality. 
Furthermore, the layer~1 attention directs over 98\% of its weight to the true stack-top position in all variants, compared to a uniform baseline of 1.3\%--1.7\% (i.e., what equal attention across all positions would give). Finally, causal ablations of the probe-aligned distance subspace yield $\Delta=0.000$ for all $\alpha$ across all variants, consistent with our base model results.

\subsection{Corruption strategy for activation patching}
\label{app:corruption}
We construct activation patching pairs by replacing a matched opening bracket with a different bracket type and evaluating at the corresponding closing bracket prediction position, only keeping pairs where the corrupted model predicts incorrectly. This strategy directly targets the bracket-type retrieval mechanism that layer~1 attention implements.

\subsection{Activation patching: full offset heatmap}
\label{app:patching-full}
\begin{figure}[H]
    \centering
    \includegraphics[width=\linewidth]{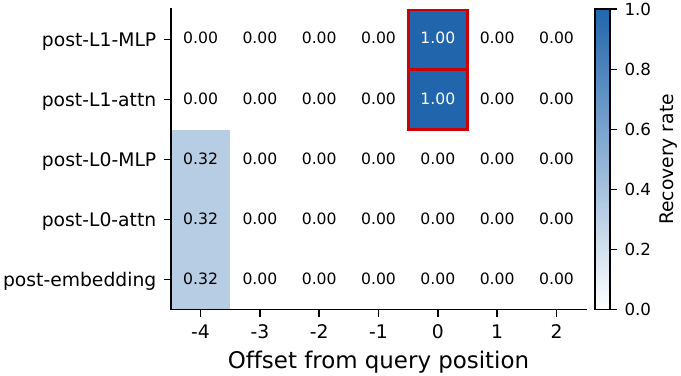}
    \caption{Full activation patching recovery heatmap across cache points (rows) and offsets from the closing bracket (columns).}
    \label{fig:patching-full}
\end{figure}

The main-text patching figure reports recovery rate only at the closing-bracket query position. Figure~\ref{fig:patching-full} shows a larger offset sweep, illustrating that recovery is concentrated at offset $0$ for post-layer-1 cache points, with negligible recovery elsewhere. This confirms that the critical retrieval step is localized to a single sublayer and position.

\subsection{Distance-sweep numeric summary}
\label{app:distance-sweep}

At low ranks ($r \in \{2, 4, 8\}$), the probe-aligned subspace is behaviorally inert while random rank-matched ablation causes substantial drops, showing the decoded subspace is not causally privileged. As rank grows, both ablations converge to near-total accuracy loss; at intermediate ranks ($r \le 32$) the probe-aligned subspace remains substantially less damaging than random. This rules out the possibility that distance is used but distributed across too many dimensions for the low-rank probe to capture, further reinforcing that the decodable distance geometry does not drive the model's predictions. \footnote{At rank $r=d=64$, the ablation is degenerate, since any orthonormal basis spans the full residual space. In this case, $P_r = I$, and the probe-aligned and random ablations both zero the residual stream identically. We report this row only for completeness, and treat its values as noise.}

\begin{table}[!ht]
\centering
\small
\setlength{\tabcolsep}{3pt}
\caption{Distance-subspace ablations across ranks ($d=64$, IID). Probe $\Delta$ = accuracy change when removing the probe-aligned subspace; Rand $\Delta$ = same for a random rank-matched subspace; Gap = probe minus random. Rank 64 is aggregated across five model configurations ($d=64$ base + four harder variants) with $\alpha=1.0$ and five random subspace seeds per model; ranks 2--32 are from the original rank-sweep on the base model.}
\label{tab:distance-sweep}
\begin{tabular}{cccc}
\toprule
Rank & Probe $\Delta$ & Rand $\Delta$ & Gap \\
\midrule
2  & $-0.000\pm0.000$ & $-0.197\pm0.169$ & $+0.197$ \\
4  & $-0.006\pm0.017$ & $-0.301\pm0.185$ & $+0.295$ \\
8  & $-0.015\pm0.028$ & $-0.485\pm0.258$ & $+0.470$ \\
16 & $-0.316\pm0.447$ & $-0.625\pm0.262$ & $+0.309$ \\
32 & $-0.458\pm0.458$ & $-0.751\pm0.220$ & $+0.293$ \\
64 & $-0.931\pm0.024$ & $-0.959\pm0.018$ & $+0.028$ \\
\bottomrule
\end{tabular}
\end{table}

\begin{figure}[H]
    \centering
    \includegraphics[width=\linewidth]{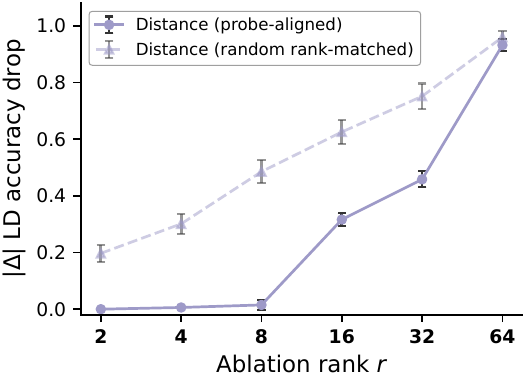}
    \caption{Accuracy drop from ablating the probe-aligned distance subspace (solid) vs.\ a random subspace of equal rank (dashed), across ranks $r \in \{2, 4, 8, 16, 32, 64\}$.}
    \label{fig:distance-sweep}
\end{figure}

\subsection{$\alpha$-sweep (intervention strength)}
\label{app:alpha-sweep}
\begin{figure}[H]
    \centering
    \includegraphics[width=\linewidth]{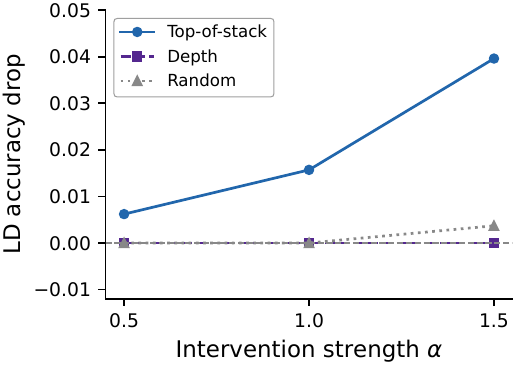}
    \caption{Residual-stream ablation at three intervention strengths.}
    \label{fig:causal_evidence}
\end{figure}

All our residual-stream ablation results in the main text use $\alpha = 1$. For completeness, we also present results of ablating with $\alpha \in \{0.5, 1.5\}$ to check robustness to intervention strength. This sweep further verifies our claim that depth and distance ablation have negligible behavioral effect, and random rank-matched ablation has a much larger effect at low ranks. Figure~\ref{fig:causal_evidence} shows the full $\alpha$-sweep.

\subsection{Query/key-space interventions}
\label{app:qk-interventions}

We additionally test whether probe-aligned directions are causally involved in the computation of attention scores. Let $q = W_Q h$ and $k = W_K h$ be the query and key vectors, and let $U_r$ be the probe-aligned residual basis. We project this $U_r$ into the query/key space via $B_Q = \mathrm{orth}(W_Q U_r)$ and $B_K = \mathrm{orth}(W_K U_r)$, then remove these components from $q$ and $k$ before recomputing the attention weights.

When doing this, we find that removing the probe-aligned components from query/key space causes large drops in stack-top attention at subspace dimensions 8, 16, and 32 (with mean changes of $-0.778\pm0.082$, $-0.816\pm0.041$, and
$-0.890\pm0.059$), which are much larger than the weak effects seen under residual-stream ablation. Query-only and key-only removals are also both strongly negative, meaning both sides of the attention score contribute. Furthermore, retaining only the probe-aligned component is also harmful, indicating these directions are necessary but not sufficient
for the routing computation. 

This is consistent with our activation patching result: probe-aligned directions participate in layer~1 attention routing, but the full mechanism requires additional structure that the low-rank probes do not capture.

\subsection{Probe accuracy vs.\ sequence position}
\label{app:probe-position}

\begin{figure}[H]
    \centering
    \includegraphics[width=\linewidth]{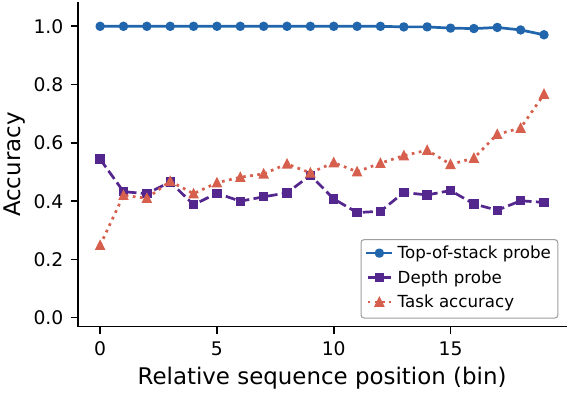}
    \caption{Probe and task accuracy vs.\ relative sequence position.}
    \label{fig:probe-position}
\end{figure}

Figure~\ref{fig:probe-position} shows our probe and task accuracy as a function of relative sequence position. The top-of-stack probe is near-perfect ($\approx 100\%$) throughout, dropping only to 97.1\% at sequence end. The depth probe Pearson $r$ hovers around 0.40--0.45 with no positional trend, consistent with weak linear decodability. Next-token accuracy rises from 25\% to 77\% near sequence end, reflecting increasing predictability as sequences approach termination.

\subsection{Probe corruption tracking}
\label{app:probe-corruption}

\begin{figure}[H]
    \centering
    \includegraphics[width=0.75\linewidth]{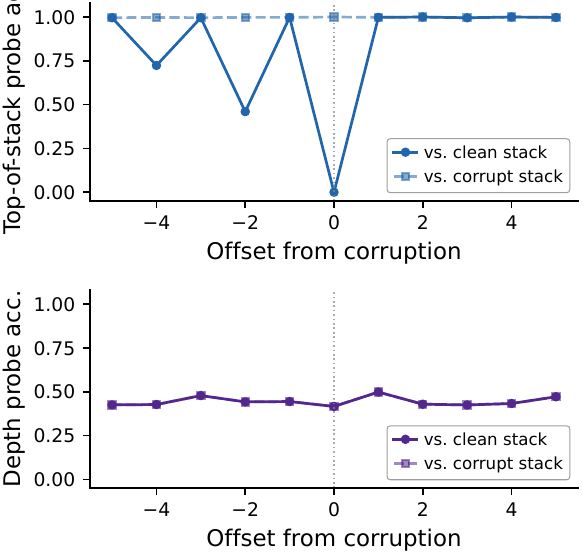}
    \caption{Probe tracking on corrupted inputs. Top: top-of-stack probe accuracy versus clean and corrupt ground truth, measured by offset from the corruption point. Bottom: depth probe accuracy is unchanged by the corruption, confirming that bracket-type corruption does not affect depth.}
    \label{fig:corruption}
\end{figure}

Figure~\ref{fig:corruption} shows probe accuracy on corrupted inputs by offset from the corruption point. The top-of-stack probe tracks the corrupted input faithfully (100\% vs.\ corrupt stack, 0\% vs.\ clean stack at offset~0), while the depth probe is flat and identical for both comparisons across all offsets, confirming that our corruption procedure correctly isolates the type-tracking mechanism.

\subsection{Multi-seed replication}
\label{app:multiseed}

We train three additional seeds (7, 11, 22) on the base Dyck-$(20,10)$, $d=64$ model using 150k steps, and two additional seeds (11, 22) on the Dyck-$(40,18)$ harder variant. We find that seeds 7, 11, 22, and 42 consistently learn the same mechanism, with mean layer~1 stack-top attention mass $\approx 0.97$, edge-masking yielding $\Delta \approx-0.938 \pm 0.003$, and activation patching recovery jumping from 0 to 1 at post-layer-1-attention (logit-diff recovery $1.000 \pm 0.000$). We note that the 1.000 values are an architectural identity, since once our clean post-layer-1 attention state is restored at the query position, all remaining computation deterministically reproduces the clean output.

We also train four harder model variants with different language complexities --- specifically with $(k,m)\in\{(20,18),(30,14),(40,10),(40,18)\}$ --- all using independent random
initializations. The consistency across these variants (Table~\ref{tab:harder-full}) provides evidence that our findings are not specific to a single random seed.

We note that one additional seed converged to a qualitatively different circuit --- reaching comparable long-distance accuracy with layer~0 attention specialized rather than layer~1 --- which we leave as an open question about the uniqueness of the solution rather than a failure of our discovered mechanism.

\subsection{Natural language pilot: subject-verb agreement}
\label{app:nl-pilot}

To test if our dissociation extends beyond formal languages, we train a 2-layer, 1-head transformer with the same architecture ($d=64$) on a templated subject-verb agreement task. We evaluate the three main properties identified in the Dyck setting: decodability without causality (Q1), causal necessity for a single edge (Q2), and sublayer localization (Q3).

In our setup, each sentence follows a controlled template with a single unambiguous subject, zero or more prepositional phrase attractors with independently randomized numbers, and a dedicated \texttt{<verb>} query token at the end. This design eliminates the redundant syntactic cues present in unconstrained English, forcing our model to retrieve the subject number via attention rather than local token identity.

Across four seeds (42, 7, 11, 22), our model achieves 99.2--99.4\%
accuracy uniformly across attractor counts (0--3), confirming no
shortcut exploitation. Furthermore, Q1, Q2, and Q3 all hold: the subject number probe accuracy remains $\geq 0.99$ IID and OOD, while the attractor count probe $R^2$ collapses from positive IID to $-12.3$ OOD, and activation patching localizes the computation to layer~1 attention (logit-diff recovery $1.000$ at post-layer-1 attention, $0.000$ at post-embedding), replicating our Dyck localization result exactly.

For Q2, seeds 42, 7, and 11 show a catastrophic accuracy drop ($\approx 0.50$) when the single subject-to-verb attention edge at layer~1 is masked. Interestingly, however, seed~22 only routes subject information partially through layer~0 attention, so simply masking layer~1 is insufficient. Nevertheless, jointly masking layer~0 and layer~1 is catastrophic, confirming that our mechanism concentrates in a single attention layer. This behavior mirrors our outlier seed discussed in Appendix~\ref{app:multiseed}, suggesting that solution multiplicity across layers is a general phenomenon rather than a Dyck-specific artifact.

In summary, all three findings replicate in natural language: hierarchical variables are decodable but not causally necessary (Q1), a single attention edge is causally necessary for task performance (Q2), and critical computation localizes to a single sublayer (Q3).
\end{document}